\documentclass[10pt,twocolumn,letterpaper]{article}

\usepackage[pagenumbers]{iccv} % To force page numbers, e.g. for an arXiv version

\definecolor{iccvblue}{rgb}{0.21,0.49,0.74}
\usepackage[pagebackref,breaklinks,colorlinks,allcolors=iccvblue]{hyperref}

\title{Crafting Query-Aware Selective Attention for Single Image Super-Resolution}

\author{
Junyoung Kim$^{1}$,\enspace Youngrok Kim$^{2}$,\enspace Siyeol Jung$^{3}$,\enspace Donghyun Min$^{4}$ \\
	\textsuperscript{1}POSTECH,\enspace \textsuperscript{2}Kyunghee University,\enspace \textsuperscript{3}UNIST,\enspace \textsuperscript{4}Sogang University\\
{\tt\small junyoungkim@postech.ac.kr, faller825@khu.ac.kr,} \\ 
{\tt\small siyeol@unist.ac.kr, mdh38112@sogang.ac.kr}
}

\newcommand{\proposed}{SSCAN}
\newcommand{\fgca}{fine-grained context-aware attention}
\newcommand{\swinir}{SwinIR}
\newcommand{\maxvit}{MaxVit}
\newcommand{\biformer}{BiFormer}

\usepackage{multirow}
\usepackage{array}
\usepackage{makecell}

\begin{document}
\maketitle
\begin{abstract}
Single Image Super-Resolution (SISR) reconstructs high-resolution images from low-resolution inputs, enhancing image details. While Vision Transformer (ViT)-based models improve SISR by capturing long-range dependencies, they suffer from quadratic computational costs or employ selective attention mechanisms that do not explicitly focus on query-relevant regions. Despite these advancements, prior work has overlooked how selective attention mechanisms should be effectively designed for SISR.
We propose \proposed{}, which dynamically selects the most relevant key-value windows based on query similarity, ensuring focused feature extraction while maintaining efficiency. In contrast to prior approaches that apply attention globally or heuristically, our method introduces a query-aware window selection strategy that better aligns attention computation with important image regions.
By incorporating fixed-sized windows, \proposed{} reduces memory usage and enforces linear token-to-token complexity, making it scalable for large images. Our experiments demonstrate that \proposed{} outperforms existing attention-based SISR methods, achieving up to 0.14 dB PSNR improvement on urban datasets, guaranteeing both computational efficiency and reconstruction quality in SISR.
\end{abstract}    
\section{Introduction}
\label{sec:introduction}
Single Image Super-Resolution (SISR) aims to reconstruct high-resolution (HR) images from low-resolution (LR) inputs, enhancing image quality and detail preservation in various applications.
SISR is widely used in fields such as satellite imagery and digital photography, where high-resolution outputs are crucial for accurate interpretation and analysis~\cite{khurana2023natural, fu2023image, liu2021splitsr, conde2023efficient, tian2022generative, lee2019mobisr}.
Even for large-scale applications such as remote sensing and medical imaging, precise and detailed enhancement is necessary to extract meaningful information~\cite{wang2018esrgan, pathak2018efficient, xu2024videogigagan}. However, deploying SISR in real-world scenarios requires a balance between performance and efficiency, as high-resolution processing must be lightweight for practical use in real-time systems~\cite{ayazoglu2021extremely, gankhuyag2023lightweight}.

For instance, modern surveillance systems and smartphones incorporate on-device SISR processing under constrained memory environments, ensuring real-time performance~\cite{aakerberg2022real, cabrera2021toward}.
Similarly, social media platforms such as Instagram and Snapchat use real-time super-resolution to enhance image quality while maintaining computational efficiency on mobile devices~\cite{kinli2021instagram}. These examples highlight the necessity of designing SISR models that achieve high-quality image restoration without excessive computational cost.

Recent deep learning-based SISR methods have significantly improved performance. Convolutional Neural Networks (CNNs) have been widely adopted due to their ability to extract local image features effectively~\cite{dong2014learning, dong2016accelerating, lim2017enhanced, kim2016accurate}.
However, CNNs struggle to model long-range dependency due to their inherent locality inductive bias~\cite{dosovitskiy2020image}, making them less suitable for realistic human-interactive environments.
To address this limitation, Transformer-based architectures incorporating self-attention mechanisms have been developed~\cite{vaswani2017attention}.
These models dynamically weigh the importance of different regions in the input sequence, enabling better global feature learning~\cite{vaswani2017attention, dosovitskiy2020image, chen2020pre, liang2021swinir, Mei_2021_CVPR, chen2023activating}.
Despite their advantages, a key challenge with typical transformer-based SISR techniques is their high computational complexity. The typical self-attention mechanisms calculate token affinity across all spatial locations, causing computational costs to grow quadratically with image size. This limits their deployment in resource-constrained environments~\cite {chen2023activating}.

\begin{figure}
    \includegraphics[width=\linewidth]{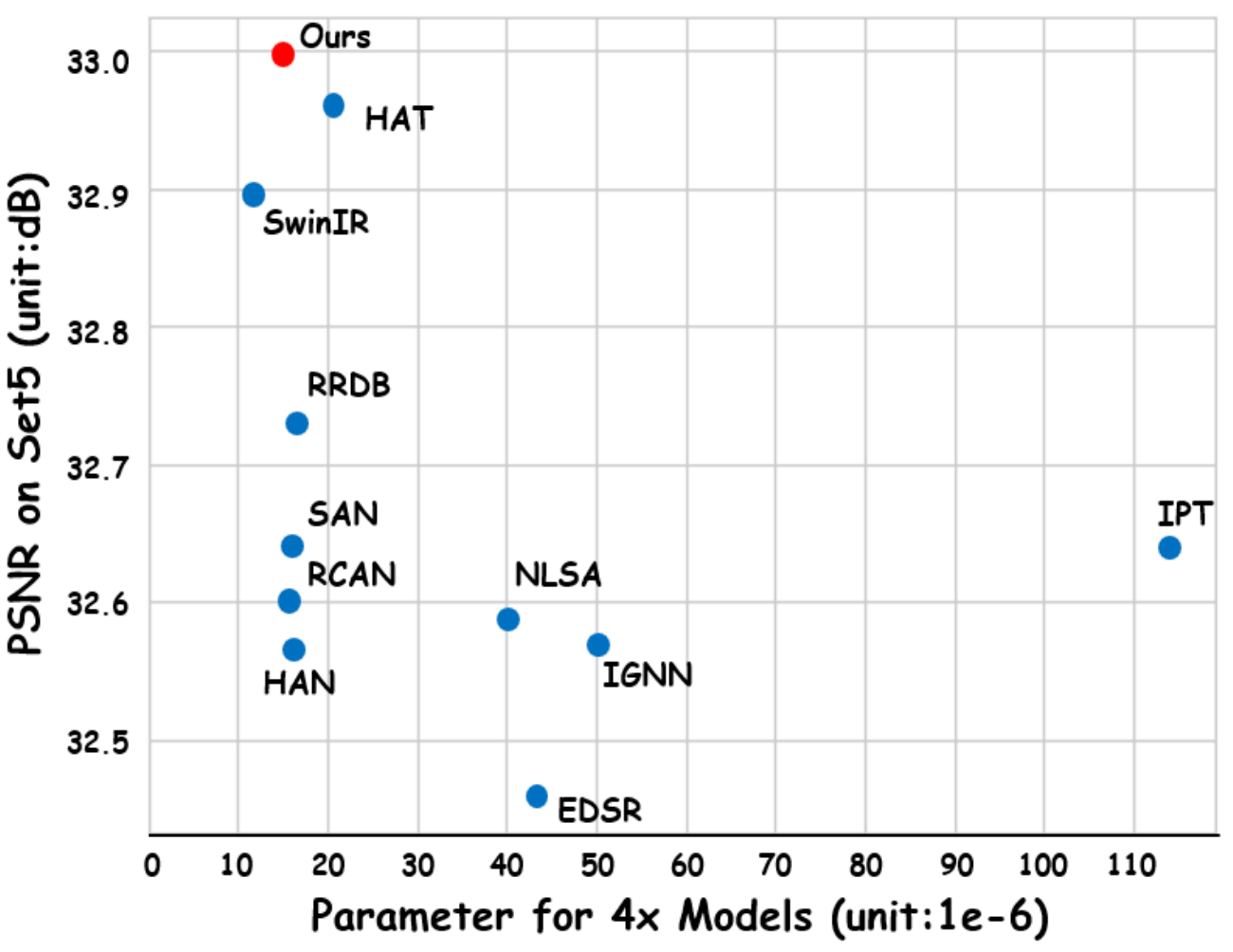}
    \caption{The comparisons of recent transformer-based SR models in terms of PSNR and parameters on Urban100 dataset. Our model (\proposed{}) outperforms the SOTA models (×4) by up to 0.14dB in PSNR.}
    \label{fig:fig1} 
\end{figure}

To mitigate these computational challenges, recent models have explored various types of selective attention mechanisms as an alternative~\cite{dong2022cswin, tu2022maxvit, wang2023crossformer, zhu2023biformer, zhang2024transformer}. For example, \swinir{} adopts a window-based self-attention approach~\cite{liu2021swin}, restricting computations to non-overlapping local windows and significantly reducing complexity.
Similarly, grouped attention methods~\cite{mei2024gtmfuse} aims to enhance efficiency by selectively processing image regions.
Although these approaches improve efficiency, they still rely on fixed partitioning rather than dynamically selecting relevant regions, limiting their ability to adapt to varying semantic importance.
Consequently, they struggle to fully capture the contextual relevance of image regions.

% To the best of our knowledge, no prior work has architecturally investigated how selective attention should be applied and designed to the SISR task.
In this work, we try to bridge this gap by proposing \textbf{\proposed{}} (\textbf{\underline{S}}uper-resolution using \textbf{\underline{S}}elective and \textbf{\underline{C}}ontext-aware \textbf{\underline{A}}ttention-based \textbf{\underline{N}}etwork).
\proposed{} dynamically selects the most relevant regions based on query-key similarity tailored for super-resolution.
This context-aware selective attention ensures that only the most informative image regions contribute to reconstruction, reducing computational complexity even for large-scale images while enhancing reconstruction fidelity.
Our main contributions are the following:

\begin{enumerate}
    \item We proposed \fgca{} (FGCA), selective top-k attention for SISR that dynamically attends to the most relevant semantic regions, regardless of image size.

    \item \proposed{} maintains the same architecture and hyperparameters as \swinir{}, with the only modification being the integration of our FGCA block into the attention module.
    
    \item \proposed{} significantly outperforms \swinir{}, achieving up to $\textbf{0.14dB}$ improvement over the state-of-the-art selective attention-based SISR model (Refer to Figure~\ref{fig:fig1}).

    \item Despite its performance, \proposed{} maintains computational efficiency, not exceeding the quadratic overhead increase observed in the recent selective attention.
\end{enumerate}

\section{Related Work}
\label{sec:related}
\subsection{Image Super-resolution}
Image super-resolution (SR) is designed to enhance the visual quality and details of low-resolution (LR) images by converting them into their high-resolution (HR) equivalents~\cite{anwar2020deep}.
This task is fundamentally an ill-posed inverse problem, where multiple HR solutions could theoretically exist for a single LR input.
This complexity increases with the scaling factor, emphasizing the need for sophisticated approaches that can infer accurate HR images.
Traditionally, SR techniques have modeled the relationship between LR and HR images through various degradation functions.
Typically, these functions include bicubic interpolation, which is often augmented with noise and other custom kernels to mimic real-world conditions.
The selection of a particular degradation function significantly influences the performance of SR models, as it dictates the initial assumptions about how the LR image degrades.

Over the last few decades, deep learning has revolutionized SR by learning from extensive prior data~\cite{dong2014learning, dong2016accelerating, lim2017enhanced, kim2016accurate, hui2019lightweight}.
Beginning with models such as SRCNN~\cite{dong2014learning} that utilized deep convolutional neural networks, the field has experienced a notable shift towards increasingly complex architectures. These are designed to more effectively capture and reconstruct fine image details.
Adversarial learning further advanced the field by emphasizing perceptual quality, thereby enhancing the realism of upsampled images~\cite{ledig2017photo, wang2018esrgan, rakotonirina2020esrgan+}.

\begin{figure*}[ht]
    \centerline{\includegraphics[width=\linewidth]{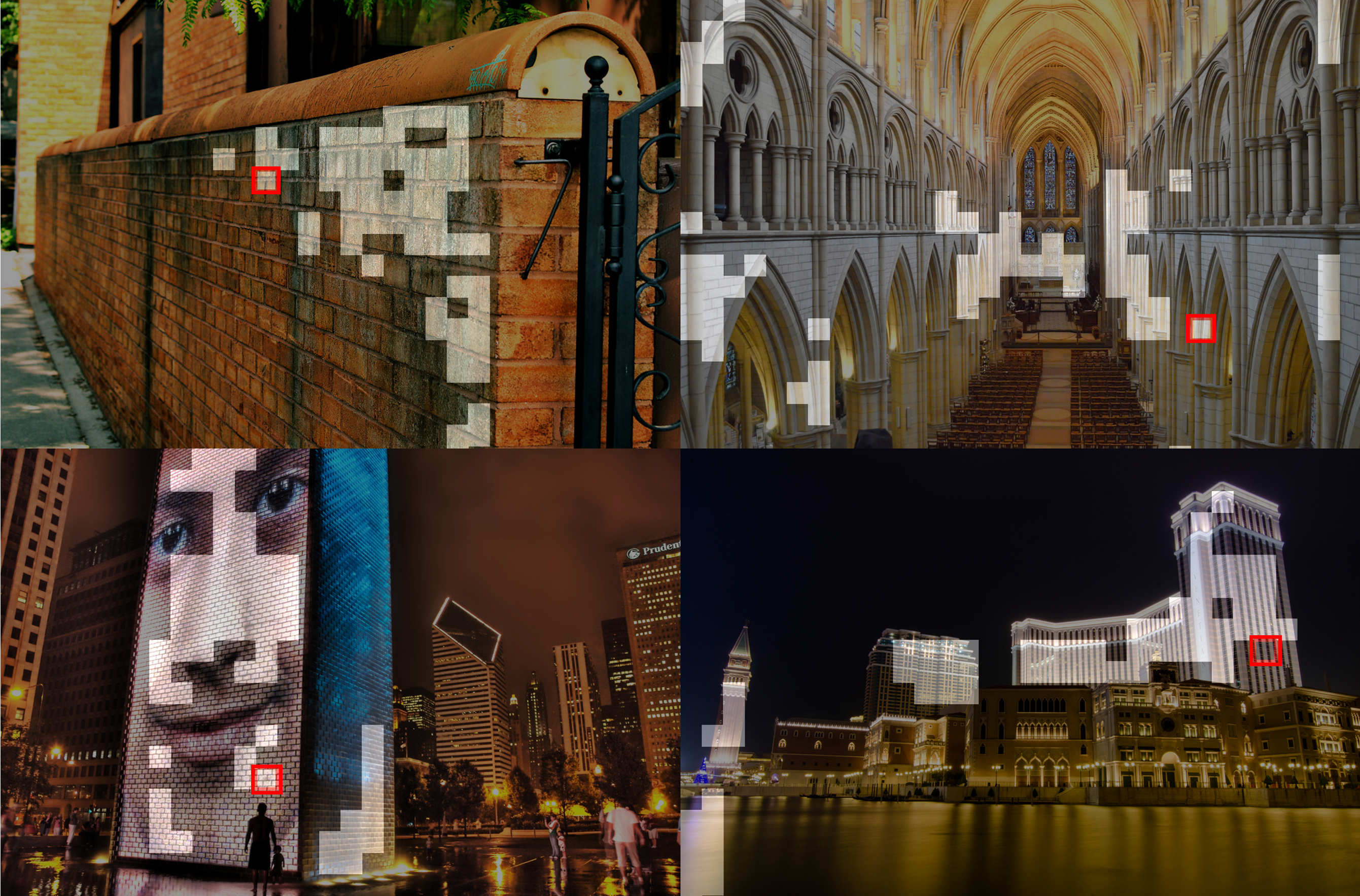}}
    \caption{Visualization of our proposed Fine-Grained Context-Aware Attention (FGCA) on Urban100 dataset: Red boxes indicate query regions, while white boxes denote corresponding key regions.}
    \label{sim_fig1}
    \vspace{-0.15in}
\end{figure*}

\subsection{Selective Attention based Transformers in Super-resolution}

The Transformer architecture~\cite{vaswani2017attention}, originally developed for natural language processing, has been successfully adapted to vision tasks, including super-resolution (SISR).
The Vision Transformer (ViT)~\cite{dosovitskiy2020image} applies self-attention to image patches, treating each as a token and modeling their global interactions.
While this approach effectively captures long-range dependencies, it generates low-resolution feature maps that compromise the capture of fine-grained spatial details.
Additionally, its global attention mechanism scales quadratically with input size, resulting in significant computational overhead.

To address these issues, selective attention-based SISR models have been introduced to improve computational efficiency while preserving or enhancing key spatial details. The selective attention methods dynamically determine which spatial regions or feature representations to attend to, rather than processing all token pairs indiscriminately.
This general paradigm can be further categorized into several key branches: local, axial, and dynamic attention, each offering distinct advantages in balancing efficiency and feature expressiveness.

\textbf{Local attention (regionally constrained selective attention).}
Local attention is one of the earliest forms of selective attention, where models restrict attention computations to localized regions to reduce complexity while preserving fine details. Swin Transformer~\cite{liu2021swin} employs a hierarchical model using window-based self-attention, limiting attention operations to non-overlapping local windows.
This method reduces computational costs significantly while maintaining spatial precision.
Moreover, shifted window mechanisms [36, 24] help extend contextual understanding by allowing gradual information exchange between neighboring regions. While this class of attention~\cite{yang2021focal, chen2023activating} can dramatically 
reduce the computational burden and improve the model's capacity to incorporate detailed spatial information across various scales.

\textbf{Axial attention (structured selective attention).}
Axial attention builds upon local attention by further constraining self-attention operations along specific spatial axes (rows and columns) rather than over entire feature maps.
For example, \maxvit{}~\cite{tu2022maxvit} introduces an axial-based attention mechanism that processes images in structured blocks and grids, enabling a balance between local detail capture and global contextual understanding.
Although this method offers computational efficiency compared to fully global attention, its structured nature may impose rigid constraints on adaptability, making it less effective in handling non-uniform feature distributions across images.

\textbf{Dynamic attention (adaptive selective attention).}
Dynamic attention, also called adaptive selective attention, provides a more flexible approach than conventional selective attention mechanisms. Instead of relying on predefined local or axial partitions, dynamic attention allows adaptively adjusting their attention structures based on input characteristics. Recent studies~\cite{zhu2023biformer} have integrated dynamic attention mechanisms into vision transformers for selectively refining which regions contribute most to feature representation.
This adaptability significantly improves both computational efficiency and the ability to capture complex spatial structures~\cite{wang2021not,tang2022quadtree,xie2022clustr}.
However, existing dynamic attention mechanisms have been primarily designed for general vision tasks. In the context of SISR, a specialized design is required to guarantee fine texture restoration, edge preservation, and repetitive pattern processing.

\section{Proposed Method}
\label{sec:method}

\begin{figure*}[ht]
    \begin{center}
    \includegraphics[width=0.9\linewidth]{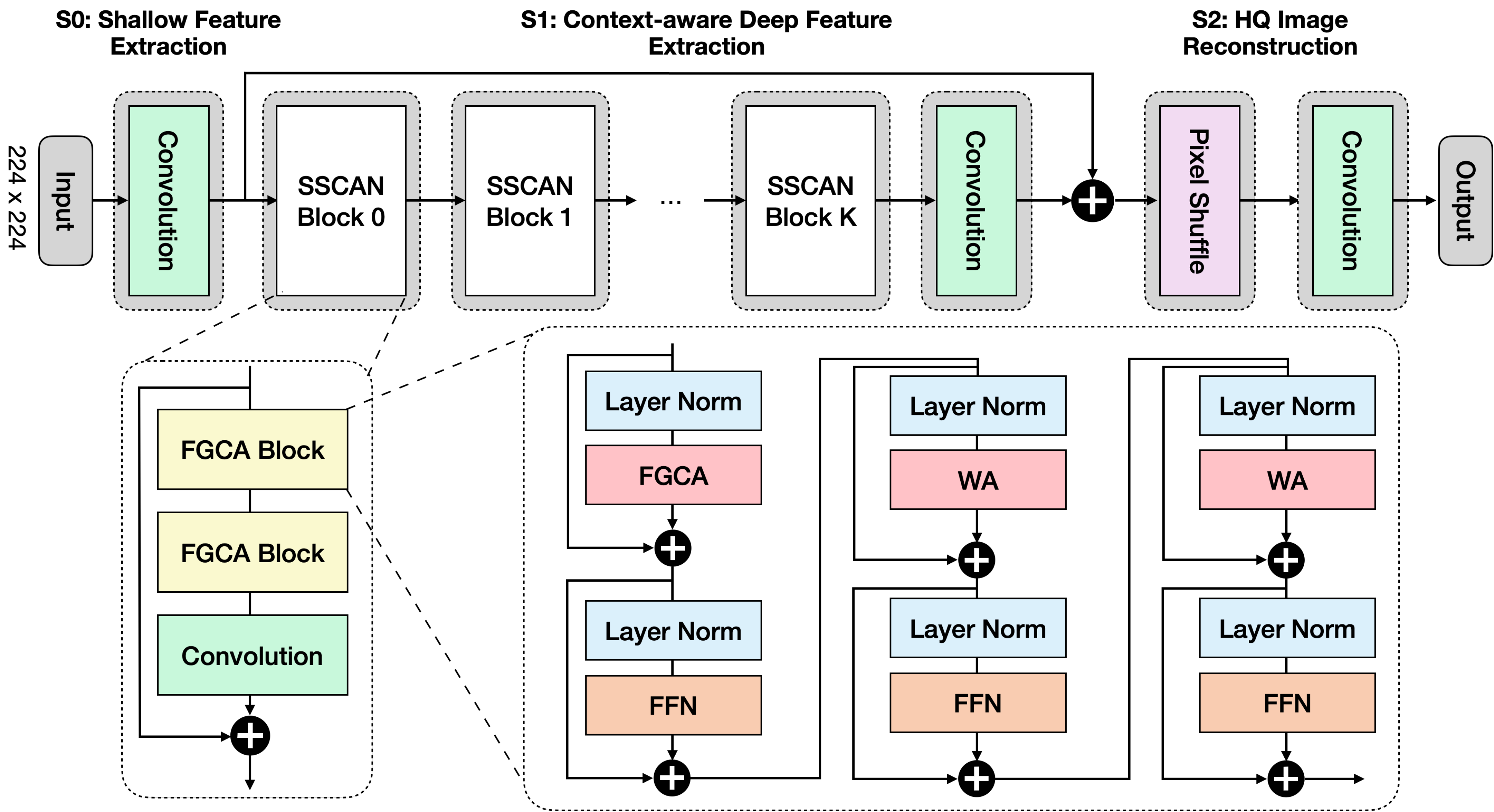}
    \end{center}
    \caption{The overall architecture of \proposed{} model and the composition of \proposed{} block. Each \proposed{} block has two FGCA blocks in this illustration.
    }
    \label{fig:overview}
\end{figure*}

Our study establishes and provides acceptable architecture design that adopts query-aware and adaptive selective attention for SISR.
In this section, we present an overview of our \proposed{} model, as illustrated in Figure~\ref{fig:overview}.
We then provide a detailed description of our \fgca{} (FGCA) for SISR and introduce core component block, namely the Residual \proposed{} Block (SSCAN block).

\subsection{Network architecture}
As shown in Figure~\ref{fig:overview}, \proposed{} consists of 3-stages.
For a low-resolution image $I\! _{LR}\in \mathbb{R}^{H \times W \times C_{in}}$, where ${H, W, C_{in}}$ denote the height, width, and channels of the input image, the feature map is extracted through a 3$\times$3 convolution layer $H_{\textit{SFE}}$ in the shallow feature extraction step.

\begin{equation}
F_{0}=H_{\textit{SFE}}(I\! _{LR}),
\end{equation} 

The subscript zero in $F_{0}$ denotes the feature obtained after the shallow feature extraction stage.
We stacked multiple \proposed{} blocks in the deep feature extraction part to extract deeper features.  

\begin{equation}
F_{DF}=H_{DF}(F\! _{0}),
\end{equation} 

where $F\! _{DF}\in \mathbb{R}^{H \times  W \times  C}$ is the extracted deep feature and $H_{\textit{DF}}(\cdot)$ is the deep feature extraction module.

After deep feature extraction, we use the residual path to reflect the features extracted by shallow feature extraction. $H_{\textit{Conv}}$ is the last convolution layer in the deep feature extraction process and $K$ is the number of SCNB blocks.
\begin{equation}
F_{DF}=H_{\textit{Conv}}(F\! _{K}) + F_{0},
\end{equation} 
In the reconstruction part, we reconstruct the features that have passed through the model into high resolution features using the reconstruction module consisting of the pixel shuffle function. 
\begin{equation}
I_{SR}=H_{\textit{rec}}(F\! _{DF}),
\end{equation} 
where $I\! _{SR}\in \mathbb{R}^{H \times  W \times  C_{in}}$ is reconstructed super-resolution image, and $H_{\textit{rec}}(\cdot)$ is reconstruction function to upscale $I\! _{SR}$ to proposed scale.

\subsection{Design of Residual \proposed{} Block}

Our model consists of multiple stacks of \proposed{} block.
It contains \textit{L} \fgca{} (FGCA) blocks and a single Conv(·) layer for feature transition at the last position.
Given the feature $F_{i,0}$ of input, we can extract deep features using $L$ layers of FGCA.
\begin{equation}
F_{i,j}={\textit{H}_{i,j}}(F_{i,j-1}) + F\! _{i,0},
\end{equation}
where $H_{i,j}(\cdot)$ is the j-th FGCA in the i-th FGCA.\\
FGCA block consists of FGCA to extract context-aware features, window attention (WA), and shifted window attention (SWA) to extract local features.
This structure is designed to capture both local features and context-aware features adaptive to the query.
Also, We positioned the FGCA layer ahead of the WA layer to capture global and context-aware features first to prevent potential bias between local and global features.
This led to an increase in SISR performance, as detailed in Section~\ref{sec:ablation_study}.

\subsection{Fine-grained Context-aware Attention (FGCA)}
Inspired by~\cite{zhu2023biformer}, our strategy of FGCA is to find the similarity between the query and key-value and perform attention using only the top-k similar key-value windows through the projection, routing, and token-to-token attention steps.
In the projection step, an input image feature $\mathbf{X} \in \mathbb{R}^{H \times W \times C}$ are divided into $\mathbf{X}^r \in \mathbb{R}^{M^2 \times \frac{HW}{M^2} \times C}$ regions such that each region has $M\times{M}$ feature vectors, where $M$ is the window size. For the partitioned region, we can drive query, key and value, $\mathbf{Q, K, V} \in \mathbb{R}^{M^2 \times \frac{HW}{M^2} \times C}$ by using linear projections
In the routing step, \fgca{} gathers the top-k similar key-value windows.
The relevant windows can be identified by using Equation~\ref{eq:routing_eq}, which computes the relevance between the two regions.

\begin{equation}
    \mathbf{A}^r = \mathbf{Q}^r (\mathbf{K}^r)^T
    \label{eq:routing_eq}
\end{equation}

where $\mathbf{Q}^r, \mathbf{K}^r \in \mathbb{R}^{M^2 \times C}$ denote that the query and key region.
Entries in the adjacency matrix $\mathbf{A}^r$ indicate how semantically related the query and key are.
Only the top-k image regions with the highest similarity are routed to the token-to-token attention step, while the rest row-wise are pruned.
In the token-to-token attention step, the actual attention operation is performed using Equation~\ref{eq:token-to-token_eq} on each query with the top-k selected key-value pairs.
\begin{equation}
    \mathbf{O} = \mathbf{softmax} \left(\frac{\mathbf{Q}\mathbf{K}^T}{\sqrt{\mathbf{d}}}\right) \mathbf{V}
    \label{eq:token-to-token_eq}
\end{equation}

where the $\mathbf{d}$ denotes the hidden dimensions of $\mathbf{Q}$ and $\mathbf{K}$ vectors. 
% The proposed \fgca{} distinguishes itself from previous implementations~\cite{zhu2023biformer} in terms of the window size.

In the previous works~\cite{zhu2023biformer}, the window size is variable, causing inconsistencies in image interpretation between the training and inference phases.
This discrepancy leads to mismatches in the amount of information processed, potentially driving the model towards a suboptimal solution.
On the other hand, our proposed methods maintain a consistent window size across both training and inference. It ensures stability in information processing and ensures that the intended training content is accurately reflected during inference.

Furthermore, this is expected to be more efficient in terms of computation and memory footprint as the input size increases.
According to Equation~\ref{bra_flops} and Equation~\ref{ours_flops}, our attention mechanism is more advantageous than previous methods~\cite{zhu2023biformer}, reducing the substantial load of computation during the token-to-token attention from quadratic to linear, preventing computational overload as image size increases.

\begin{equation}
    FLOPS_{Prev} 
    =  \overbrace{2 (S^2)^2 C}^\text{Routing} + \underbrace{2k {(HW)^2}{S^2}C}_{\textcolor{blue}{Attention}},
    \label{bra_flops}
\end{equation}

\begin{equation}
     FLOPS_{Ours} \\
    = \overbrace{2 {{(HW)^2}{M^4}}C}^\text{Routing} + \underbrace{2kM^2(HW)C}_{\textcolor{blue}{Attention}},
    \label{ours_flops}
\end{equation}

where $k$ is top-k parameters, $S^2$ is the variable window size proposed in~\cite{zhu2023biformer}, and $M^2$ denotes \proposed{}’s fixed window size.

\begin{figure}[t]
    \centering
    \includegraphics[width=\columnwidth]{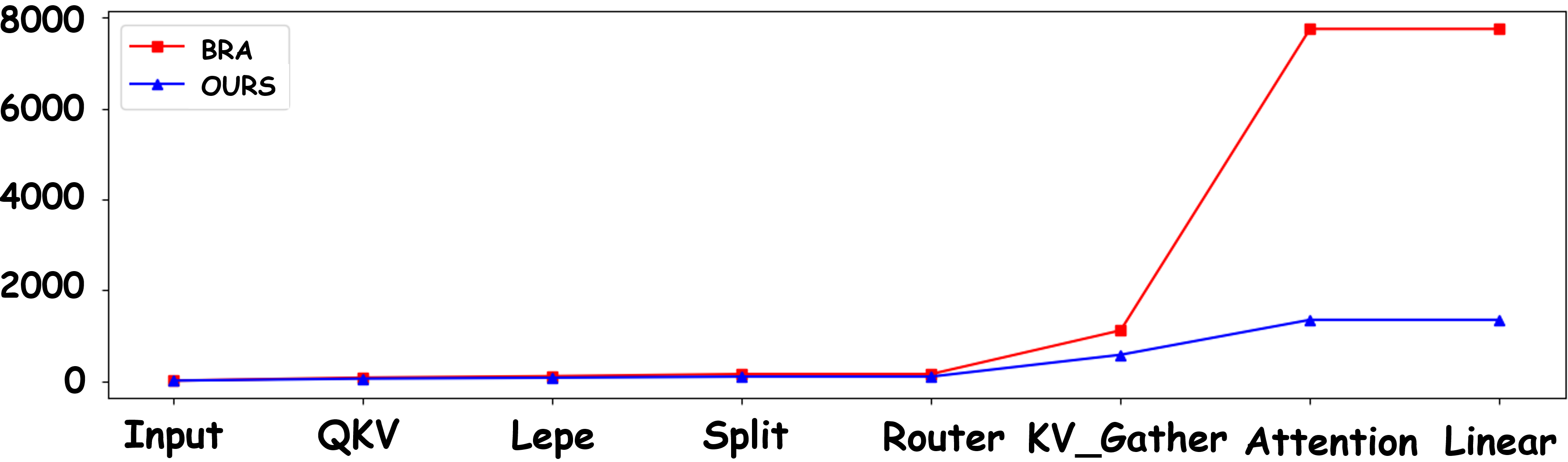}
    \label{fig:memory_flow}
    \vspace{-5mm}
    \caption{Comparison of memory consumption between \biformer{}'s attention (BRA) and our \fgca{} during calculating attention score.} 
    \label{fig:memory_expriments}
    \vspace{-3mm}
\end{figure}

For large images ($H$ and $W \gg M$), the relationship $S^2 \gg M^2$ holds, leading to significantly higher memory usage for previous works compared to \proposed{}. The experimental results shown in Figure~\ref{fig:memory_expriments} empirically validate this theoretical analysis. In Table~\ref{abl_window_size}, the result supports the theoretical advantages of \fgca{}. In particular, the better performance of smaller window sizes (4×4) on the Urban100 dataset demonstrates the effectiveness of \fgca{} in recognizing repetitive patterns. Table~\ref{order_architecture} shows that capturing global context (\fgca{}) first and refining local details (WA) is more effective for SISR, which aligns with our theoretical framework. Lastly, Figure~\ref{visual_comparison} demonstrates that \fgca{} can more accurately restore repetitive patterns and fine details, due to its ability to recognize semantic similarities.
\section{Experiments}
\label{sec:experiments}

\subsection{Experimental Setups}
\label{sec:expr_details}
\begin{table*}[ht]
\centering
\caption{Quantitative comparison with state-of-the-art methods achieved for ×2/3/4 SR. The best and second best performances are in \textcolor{red}{red} and \textcolor{blue}{blue}.}
\resizebox{\linewidth}{!}{%
\begin{tabular}{|c|c|c|c|c|c|c|c|c|c|c|}
\hline
\multirow{2}{*}{Model} & \multirow{2}{*}{\#PARAMS} & \multirow{2}{*}{Scale} & \multicolumn{2}{c|}{Set5} & \multicolumn{2}{c|}{Set14} & \multicolumn{2}{c|}{BSD100} & \multicolumn{2}{c|}{Urban100} \\ 
\cline{4-11}
                          &                           &                        & PSNR   & SSIM             & PSNR   & SSIM              & PSNR   & SSIM               & PSNR   & SSIM                  \\ 
\hline
CARN~\cite{ahn2018fast}                      & 1592K                     & \multirow{11}{*}{×2}   & 37.76  & 0.959            & 33.52  & 0.9166            & 32.09  & 0.8978             & 31.92  & 0.9256                \\ 
\cline{1-2}\cline{4-11}
IMDN~\cite{Hui-IMDN-2019}                      & 694K                      &                        & 38     & 0.9605           & 33.63  & 0.9177            & 32.19  & 0.8996             & 32.17  & 0.9283                \\ 
\cline{1-2}\cline{4-11}
AWSRN-M~\cite{wang2019lightweight}                   & 1063K                     &                        & 38.04  & 0.9605           & 33.66  & 0.9181            & 32.21  & 0.9000                & 32.23  & 0.9294                \\ 
\cline{1-2}\cline{4-11}
LAPAR-A~\cite{li2021lapar}                   & 548K                      &                        & 38.01  & 0.9605           & 33.62  & 0.9183            & 32.19  & 0.8999             & 32.10   & 0.9283                \\ 
\cline{1-2}\cline{4-11}
RFDN~\cite{liu2020residual}                      & 534K                      &                        & 38.05  & 0.9606           & 33.68  & 0.9184            & 32.16  & 0.8994             & 32.12  & 0.9278                \\ 
\cline{1-2}\cline{4-11}
LaticeNet~\cite{luo2020latticenet}                 & 756K                      &                        & 38.06  & 0.9607           & 33.70   & 0.9187            & 32.20   & 0.8999             & 32.25  & 0.9288                \\ 
\cline{1-2}\cline{4-11}
AAF-L~\cite{wang2020lightweight}                     & 1363K                     &                        & 38.09  & 0.9607           & 33.78  & 0.9192            & 32.23  & 0.9002             & 32.46  & 0.9313                \\ 
\cline{1-2}\cline{4-11}
A-CubeNet~\cite{hang2020attention}                 & 1376K                     &                        & 38.12  & 0.9609           & 33.73  & 0.9191            & 32.26  & 0.9007             & 32.39  & 0.9308                \\ 
\cline{1-2}\cline{4-11}
Swin-Light~\cite{liang2021swinir}                & 878K                      &                        & 38.14  & \textcolor{blue}{\textbf{0.9611}}           & 33.86  & 0.9206            & \textcolor{blue}{\textbf{32.31}}  & \textcolor{blue}{\textbf{0.9012}}             & \textcolor{red}{\textbf{32.76}}  & \textcolor{red}{\textbf{0.934}}                 \\ 
\cline{1-2}\cline{4-11}
ELAN-Light~\cite{zhang2022efficient}                & 582K                      &                        & \textcolor{blue}{\textbf{38.17}}  & \textcolor{blue}{\textbf{0.9611}}           & \textcolor{red}{\textbf{33.94}}  & \textcolor{blue}{\textbf{0.9207}}            & 32.30   & \textcolor{blue}{\textbf{0.9012}}             & \textcolor{red}{\textbf{32.76}}  & \textcolor{red}{\textbf{0.934}}                 \\ 
\cline{1-2}\cline{4-11}
\textbf{\proposed{} (Ours)} & 911K                      &                        & \textcolor{red}{\textbf{38.27}} & \textcolor{red}{\textbf{0.9615}}           & \textcolor{blue}{\textbf{33.93}} & \textcolor{red}{\textbf{0.9211}}            & \textcolor{red}{\textbf{32.33}} & \textcolor{red}{\textbf{0.9016}}             & \textcolor{blue}{\textbf{32.71}} & \textcolor{blue}{\textbf{0.9337}}                \\ 
\hline
CARN~\cite{ahn2018fast}                      & 1592K                     & \multirow{11}{*}{×3}   & 34.29  & 0.9255           & 30.29  & 0.8407            & 29.06  & 0.8034             & 28.06  & 0.8493                \\ 
\cline{1-2}\cline{4-11}
IMDN~\cite{Hui-IMDN-2019}                      & 703K                      &                        & 34.36  & 0.827            & 30.32  & 0.8417            & 29.09  & 0.8046             & 28.17  & 0.8519                \\ 
\cline{1-2}\cline{4-11}
AWSRN-M~\cite{wang2019lightweight}                   & 1143K                     &                        & 34.42  & 0.9275           & 30.32  & 0.8419            & 29.13  & 0.8059             & 28.26  & 0.8545                \\ 
\cline{1-2}\cline{4-11}
LAPAR-A~\cite{li2021lapar}                   & 544K                      &                        & 34.36  & 0.9267           & 30.34  & 0.8421            & 29.11  & 0.8054             & 28.15  & 0.8523                \\ 
\cline{1-2}\cline{4-11}
RFDN~\cite{liu2020residual}                      & 541K                      &                        & 34.41  & 0.9273           & 30.34  & 0.842             & 29.09  & 0.805              & 28.21  & 0.8525                \\ 
\cline{1-2}\cline{4-11}
LaticeNet~\cite{luo2020latticenet}                 & 765K                      &                        & 34.40   & 0.9272           & 30.32  & 0.8416            & 29.10   & 0.8049             & 28.19  & 0.8513                \\ 
\cline{1-2}\cline{4-11}
AAF-L~\cite{wang2020lightweight}                     & 1367K                     &                        & 34.53  & 0.9281           & 30.45  & 0.8441            & 29.17  & 0.8068             & 28.38  & 0.8568                \\ 
\cline{1-2}\cline{4-11}
A-CubeNet~\cite{hang2020attention}                 & 1561K                     &                        & 34.54  & 0.9283           & 30.41  & 0.8436            & 29.14  & 0.8062             & 28.40   & 0.8574                \\ 
\cline{1-2}\cline{4-11}
Swin-Light~\cite{liang2021swinir}                & 886K                      &                        & 34.62  & \textcolor{blue}{\textbf{0.9289}}           & 30.54  & \textcolor{blue}{\textbf{0.8463}}            & 29.20   & \textcolor{blue}{\textbf{0.8082}}             & 28.66  & \textcolor{blue}{\textbf{0.8624}}                \\ 
\cline{1-2}\cline{4-11}
ELAN-Light~\cite{zhang2022efficient}                & 590K                      &                        & \textcolor{red}{\textbf{34.64}}  & 0.9288           & \textcolor{blue}{\textbf{30.55}}  & \textcolor{blue}{\textbf{0.8463}}            & \textcolor{blue}{\textbf{29.21}}  & 0.8081             & \textcolor{blue}{\textbf{28.69}}  & \textcolor{blue}{\textbf{0.8624}}                \\ 
\cline{1-2}\cline{4-11}
\textbf{\proposed{} (Ours)} & 919K                      &                        & \textcolor{blue}{\textbf{34.62}} & \textcolor{red}{\textbf{0.9292}}           & \textcolor{red}{\textbf{30.61}} & \textcolor{red}{\textbf{0.8466}}            & \textcolor{red}{\textbf{29.25}} & \textcolor{red}{\textbf{0.8095}}             & \textcolor{red}{\textbf{28.74}} & \textcolor{red}{\textbf{0.8638}}                \\ 
\hline
CARN~\cite{ahn2018fast}                      & 1592K                     & \multirow{11}{*}{×4}   & 32.13  & 0.8937           & 28.60   & 0.7806            & 27.58  & 0.7349             & 26.07  & 0.7837                \\ 
\cline{1-2}\cline{4-11}
IMDN~\cite{Hui-IMDN-2019}                      & 715K                      &                        & 32.21  & 0.8948           & 28.58  & 0.7811            & 27.56  & 0.7353             & 26.04  & 0.7838                \\ 
\cline{1-2}\cline{4-11}
AWSRN-M~\cite{wang2019lightweight}                   & 1254K                     &                        & 32.21  & 0.8954           & 28.65  & 0.7832            & 27.60   & 0.7368             & 26.15  & 0.7884                \\ 
\cline{1-2}\cline{4-11}
LAPAR-A~\cite{li2021lapar}                   & 659K                      &                        & 32.15  & 0.8944           & 28.61  & 0.7818            & 27.61  & 0.7366             & 26.14  & 0.7871                \\ 
\cline{1-2}\cline{4-11}
RFDN~\cite{liu2020residual}                      & 550K                      &                        & 32.24  & 0.8952           & 28.61  & 0.7819            & 27.57  & 0.7360              & 26.11  & 0.7858                \\ 
\cline{1-2}\cline{4-11}
LaticeNet~\cite{luo2020latticenet}                 & 777K                      &                        & 32.30   & 0.8962           & 28.68  & 0.783             & 27.62  & 0.7367             & 26.25  & 0.7873                \\ 
\cline{1-2}\cline{4-11}
AAF-L~\cite{wang2020lightweight}                     & 1374K                     &                        & 32.32  & 0.8964           & 28.67  & 0.7839            & 27.62  & 0.7379             & 26.32  & 0.7931                \\ 
\cline{1-2}\cline{4-11}
A-CubeNet~\cite{hang2020attention}                 & 1524K                     &                        & 32.32  & 0.8969           & 28.72  & 0.7847            & 27.65  & 0.7382             & 26.27  & 0.7913                \\ 
\cline{1-2}\cline{4-11}
Swin-Light~\cite{liang2021swinir}                & 897K                      &                        & \textcolor{blue}{\textbf{32.44}}  & \textcolor{blue}{\textbf{0.8976}}           & 28.77  & \textcolor{blue}{\textbf{0.7858}}            & \textcolor{blue}{\textbf{27.69}}  & \textcolor{blue}{\textbf{0.7406}}             & 26.47  & 0.7980                 \\ 
\cline{1-2}\cline{4-11}
ELAN-Light~\cite{zhang2022efficient}                & 601K                      &                        & 32.43  & 0.8975           & \textcolor{blue}{\textbf{28.78}}  & \textcolor{blue}{\textbf{0.7858}}            & \textcolor{blue}{\textbf{27.69}}  & \textcolor{blue}{\textbf{0.7406}}             & \textcolor{blue}{\textbf{26.54}}  & \textcolor{blue}{\textbf{0.7982}}                \\ 
\cline{1-2}\cline{4-11}
\textbf{\proposed{} (Ours)} & 931K                      &                        & \textcolor{red}{\textbf{32.58}}& \textcolor{red}{\textbf{0.8991}}         & \textcolor{red}{\textbf{28.88}} & \textcolor{red}{\textbf{0.7878}}            & \textcolor{red}{\textbf{27.74}} & \textcolor{red}{\textbf{0.7422}}             & \textcolor{red}{\textbf{26.65}} & \textcolor{red}{\textbf{0.8021}}               \\
\hline

\end{tabular}%
}
\label{comparison_metrics}
\vspace{-0.1in}
\end{table*}

We employed the identical model architecture and hyperparameters as Light-SwinIR~\cite{liang2021swinir, zhang2022swinfir} to evaluate the comparative efficacy of window attention versus our proposed \fgca{}.
Our setup includes two FGCA blocks and four \proposed{} blocks.
The window sizes are set to $8\times8$ and 60 embedding dimensions.
We used Adam optimizer and trained for 500$k$ iterations on the DIV2K~\cite{agustsson2017ntire} datasets with a batch size of 64 and L1 loss as the loss function.

During training, we captured information using 32 top-k, increasing to 64 top-k during inference to include more context-aware information.
% We capture the information with 32 $topk$s in the training phase and 64 top-ks in the inference phase to capture more context-aware information.
To improve generalizability, we applied data augmentation techniques including random horizontal flip, random vertical flip, rotation, and random crop.
\vspace{-1mm}

\subsection{Datasets}
\label{sec:datasets}
We train with DIV2K~\citep{agustsson2017ntire} datasets. This dataset includes 800 images for training images.
We conduct performance evaluation with four standard benchmark datasets, Set5~\cite{bevilacqua2012low}, Set14~\cite{10.1007/978-3-642-27413-8_47}, BSD100~\cite{937655} and Urban100~\cite{Huang-CVPR-2015} in terms of PSNR and SSIM metrics.

\subsection{Quantitative Comparison with Existing Super-resolution Models}
\label{sec:comparison}
In Table~\ref{comparison_metrics}, 
we compare our model with the following super-resolution methods that can operate in resource-constrained environments, demonstrating state-of-the-art performance: CARN~\cite{ahn2018fast}, IMDN~\cite{Hui-IMDN-2019}, AWSRN-M~\cite{wang2019lightweight}, LAPAR-A~\cite{li2021lapar},  RFDN~\cite{liu2020residual}, LatticeNet~\cite{luo2020latticenet}, AAF-L~\cite{wang2020lightweight}, A-CubeNet~\cite{hang2020attention}, SwinIR-light~\cite{liang2021swinir} and ELAN-light~\cite{zhang2022efficient}. 
Note that Swin-Light is a lightweight version of SwinIR~\cite{liang2021swinir} enhancing to capture local information and ELAN-Light~\cite{zhang2022efficient} introduce efficient long-range shared attention mechanism using two shift-convolution network. 
Our method achieves a significant performance improvement while maintaining a similar number of parameters in previous work. 
\begin{table}[ht]
\scriptsize
\centering
\caption{An ablation study of the effectiveness of top-k during the inference phase.}
\vspace{-2mm}
\setlength\tabcolsep{2pt}
\renewcommand{\arraystretch}{1.3}
\resizebox{\columnwidth}{!}{%
\begin{tabular}{c|c|c|c|c}
\hline
\multirow{2}{*}{Top-k} & Set5           & Set14          & BSD100         & Urban100       \\
                       & PSNR / SSIM    & PSNR / SSIM    & PSNR / SSIM    & PSNR / SSIM    \\ \hline
32                     & 32.565 / 0.8991& 28.852 / 0.7875& 27.722 / 0.7421& 26.557 / 0.7997\\ \hline
48                     & \textbf{32.576} / \textbf{0.8992}& 28.869 / 0.7877& 27.731 / \textbf{0.7422}& 26.618 / 0.8013\\ \hline
64                     & \textbf{32.576} / 0.8991& \textbf{28.879} / \textbf{0.7878}& \textbf{27.735} / \textbf{0.7422} & \textbf{26.652} / \textbf{0.8021}\\ \hline
\end{tabular}%
}
\vspace{-3mm}
\label{abl_inference_topk}
\end{table}
Moreover, at a scale factor of $\times$2 and $\times$3, we can see that there are the second-best performance achievements for some datasets. In contrast, at a scale factor of $\times$4, our \proposed{} achieves the best results with surprising improvements on all benchmark datasets. This is because even if the model routes the same top-k key-value windows for all scale factors, the input image size is the largest at scale $\times$2, followed by $\times$3 and $\times$4, so the ratio of the amount of information (M$\times k$) to the image size is smallest for $\times$2 and largest for $\times$4.
Accordingly, our \proposed{} performs in terms of PSNR with -0.05dB $\sim$ 0.1dB for $\times$2, whereas 0.05dB $\sim$ 0.14dB for $\times$4.
Our \proposed{} is ideally suitable for reconstructing images from small image sizes to larger ones with higher upscaling.

\subsection{Analysis of Memory Usage}
\label{sec:memory_usage_analysis}
We analyze the memory usage of each component in the Bi-Routing Attention (BRA) mechanism of the \biformer{}~\cite{zhu2023biformer}, as depicted in Figure~\ref{fig:memory_expriments}.
While BRA surpasses state-of-the-art in numerous downstream tasks, it exhibits substantial memory consumption.
This is primarily due to the storage requirements for the $Q\times{K}$ computation results and subsequent softmax operations, as illustrated in Figure~\ref{fig:memory_expriments}.
These intermediate results are stored in DRAM, leading to increased memory access overhead, which in turn causes system latency.
Such high memory demands are unsuitable for resource-constrained environments.
To alleviate this complexity, adopting a fixed window size approach can be effective.
When the input image size is exceeds (M$\times$S, M$\times$S), processes larger query, key, and value regions than (M, M). 
In these scenarios, storing computation results consumes more memory compared to \fgca{}, which consistently operates on M$\times$M window, irrespective of input size.
As a result, \fgca{} offers a memory usage advantage for super-resolving large images in resource-limited settings.
Furthermore, FlashAttention~\cite{dao2022flashattention} effectively addresses these problems by partitioning the query, key, and value matrices into blocks and computing them entirely within on-chip SRAM.
By applying 1) fixed window size and 2) FlashAttention, our \fgca{} can significantly reduce the memory usage in the key-value gather and token-to-token attention process, as shown in Figure~\ref{fig:memory_expriments}. In addition, we can reduce the overall amount of memory usage by approximately \textbf{5×}.

% \vspace{-1cm}

% \vspace{0.7cm}

\begin{table}[ht]
\scriptsize
\setlength\tabcolsep{2pt}
\renewcommand{\arraystretch}{1.3}
\centering
\caption{An ablation study of the order of Fine-grained Routing Attention and Window Attention.}
\vspace{-2mm}
\resizebox{\columnwidth}{!}{%
\begin{tabular}{c|c|c|c|c}
\hline
\multirow{2}{*}{Order} & Set5           & Set14          & BSD100         & Urban100       \\
                      & PSNR / SSIM    & PSNR / SSIM    & PSNR / SSIM    & PSNR / SSIM    \\ \hline
Only $WA$                & 32.44 / 0.8976 & 28.77 / 0.7858 & 27.69 / 0.7406 & 26.47 / 0.798  \\ \hline
$WA \rightarrow FGCA$               & 32.55 / 0.8987 & 28.874 / 0.7875& \textbf{27.735} / 0.7421& 26.626 / 0.8003 \\ \hline
$FGCA \rightarrow WA$                & \textbf{32.576} / \textbf{0.8991} & \textbf{28.879} / \textbf{0.7878} & \textbf{27.735} / \textbf{0.7422} & \textbf{26.652} / \textbf{0.8021} \\ \hline
\end{tabular}%
}
% \vspace{-5mm}
\label{order_architecture}
\end{table}
\subsection{Ablation Study}
\label{sec:ablation_study}
We conducted some ablation studies to prove that our \proposed{} is more effective than other methods. All experiments were trained basically with top-k 32 and top-k 64 in inference on the scale of ×4. Based on our findings, \proposed{} provides guidelines for designing SISR model architectures.

\textbf{Effectiveness of Top-k}
In Table~\ref{abl_inference_topk}, it can be observed that increasing the top-k value during the inference phase generally results in higher PSNR and SSIM values. This improvement is attributed to the ability to capture more relevant information. However, we observe that for some datasets, there are slight or no increase about some datasets. This phenomenon occurs because, for these datasets, lower top-k is already sufficient to capture the relevant key-value windows similar to the query. In the case of the Urban100~\cite{Huang-CVPR-2015} dataset, which contains a significant amount of repetitive pattern information, there are numerous windows similar to the query so that higher top-k values enable the model to capture more relative information, resulting in notable improvements on PSNR and SSIM.

\begin{table}[t]
% \scriptsize
\centering
\caption{Ablation study of different window sizes at a scale factor of $\times$2/$\times$3/$\times$4}
\vspace{-2mm}
\label{abl_window_size}

\renewcommand{\arraystretch}{1.2}
\setlength{\tabcolsep}{4pt}

\resizebox{\columnwidth}{!}{
\begin{tabular}{c|c|c|c|c|c}
\Xhline{1.5\arrayrulewidth}
\multirow{2}{*}{Window Size} & \multirow{2}{*}{Top-k} & \multirow{2}{*}{Scale} & Set14 & BSD100 & Urban100 \\
                            &                         &                         & PSNR / SSIM & PSNR / SSIM & PSNR / SSIM \\ 
\Xhline{1.5\arrayrulewidth}

8$\times$8 & 64  & \multirow{2}{*}{\centering X2}  & \textbf{33.929} / \textbf{0.9211} & \textbf{32.326} / \textbf{0.9016} & 32.708 / 0.9337 \\  
4$\times$4 & 256 &                                & 33.911 / 0.9200 & 32.322 / 0.9014 & \textbf{32.825} / \textbf{0.9342} \\  
\Xhline{1.5\arrayrulewidth}

8$\times$8 & 64  & \multirow{2}{*}{\centering X3}  & \textbf{30.610} / \textbf{0.8466} & \textbf{29.254} / \textbf{0.8095} & 28.740 / 0.8638 \\  
4$\times$4 & 256 &                                & 30.592 / 0.8463 & 29.247 / 0.8089 & \textbf{28.804} / \textbf{0.8643} \\  
\Xhline{1.5\arrayrulewidth}

8$\times$8 & 64  & \multirow{2}{*}{\centering X4}  & 28.879 / \textbf{0.7878} & 27.735 / \textbf{0.7422} & 26.652 / 0.8021 \\  
4$\times$4 & 256 &                                & \textbf{28.887} / \textbf{0.7878} & \textbf{27.739} / \textbf{0.7422} & \textbf{26.671} / \textbf{0.8027} \\  
\Xhline{1.5\arrayrulewidth}
\end{tabular}
}
\vspace{-3mm}
\end{table}

\textbf{Analysis of Different Window Sizes}
To test the performance of \fgca{} with varying window sizes,
we train this model with window sizes of 4 and 8, and each top-k set to 256 and 64 for inference, so that the models can have the same amount of information.
As shown in Table~\ref{abl_window_size}, at a scale factor of $\times$4, the performance is slightly better at a window size of 4 than 8. 

\begin{figure*}[t]
    \vspace{-0.09in}
    \centering
    \includegraphics[width=0.85\textwidth,scale=1.0]{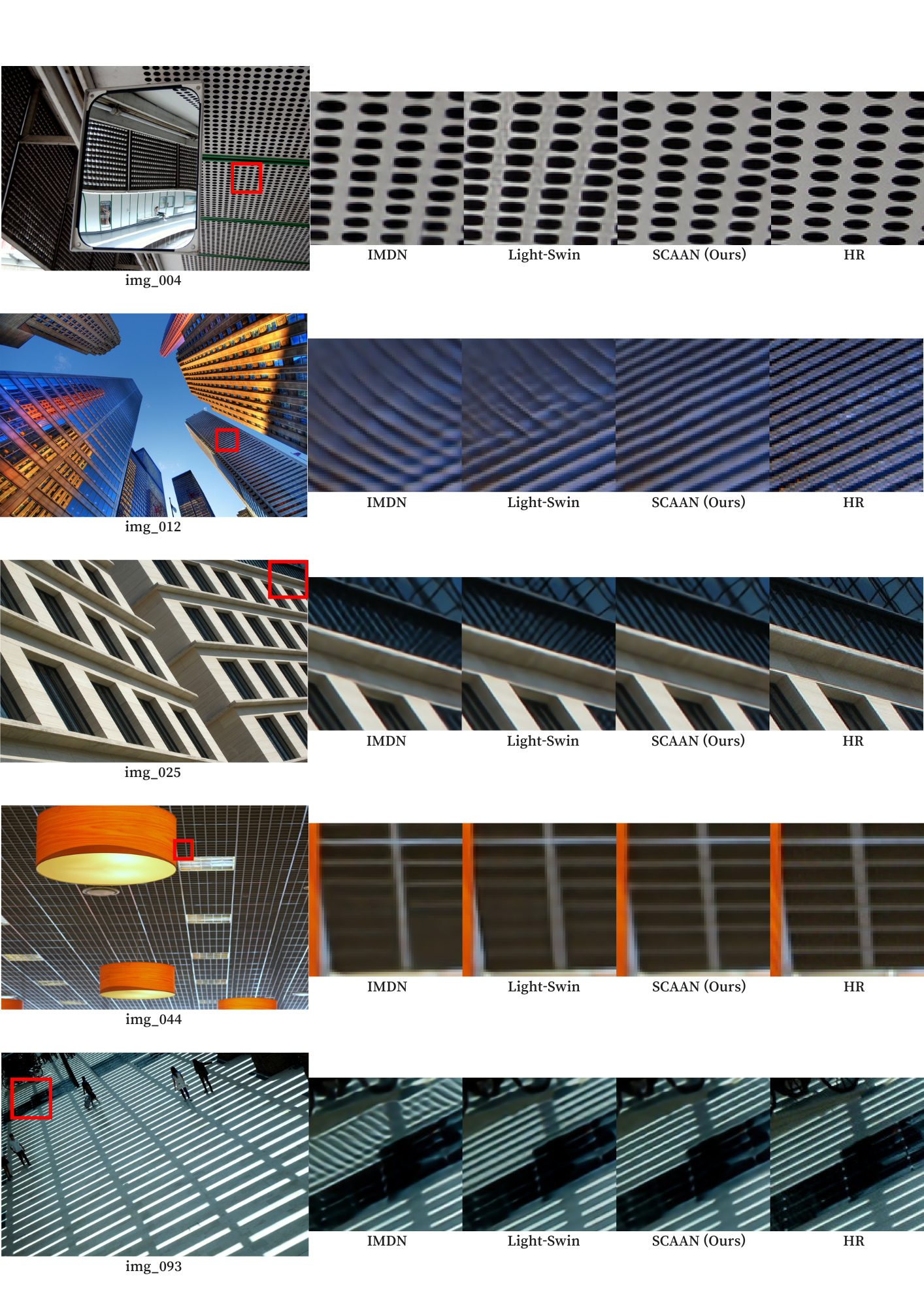}
    \caption{Visual comparisons of \proposed{} with other publicly released SISR models on Urban100 dataset (×4).}
    \label{visual_comparison} 
\end{figure*}
However, for scale factors $\times$2 and $\times$3, we find a performance drop when the window size is 4 for most datasets, but for the Urban100~\cite{Huang-CVPR-2015} dataset, there is a significant performance increase of up to 0.12dB.
Since the Urban100~\cite{Huang-CVPR-2015} dataset has more repeated patterns than other datasets, our context-aware attention mechanism reconstructs more effectively. 
% Also because the image size is larger than other datasets, our fixed window size technique allows it to capture more fine-grained information, where a large window size may unintentionally capture unrelated parts to the query, leading to a significant performance improvement over other datasets.
Also because the image size is larger than other datasets, our fixed window size technique allows it to capture more fine-grained information, where a large window size may unintentionally capture unrelated parts to the query, leading to a significant performance improvement over other datasets.
This demonstrates the importance of adapting window strategies to the inherent characteristics of the data, especially in high-resolution settings.

\textbf{Balance between Local and Global Feature}
To analyze the importance of integrating local and global features and their impact on feature extraction timing, we evaluate four different models, as shown in Table~\ref{order_architecture}.
First, Case 1 captures only local features and this cannot lead to performance improvement in the SISR tasks.
Case 2 and Case 3, which incorporate both local and global attention, exhibit substantial performance gains, confirming that integrating both feature types is beneficial for SISR tasks.

To test where to use those global features for optimal performance in super-resolution, we designed two detailed configurations: Case 2, where \fgca{} is placed before the WA layer, and Case 3, where \fgca{} is applied after the WA layer.
In Case 2, local features are extracted first, followed by global features. Conversely, in Case 3, global features are captured first, with local feature extraction occurring at the end of the block.

Experimental results indicate that a balanced integration of local and global features within a block yields the best SISR performance, rather than prioritizing one over the other. This suggests that maintaining a balanced representation of both feature types is crucial for achieving optimal results in super-resolution tasks.

\subsection{\textbf{Visual Results}}
\label{sec:visual_results}

We compare the visual quality of Urban100~\cite{Huang-CVPR-2015} SISR results using our SCAAN compared with IMDN~\cite{Hui-IMDN-2019}, SwinIR-light~\cite{liang2021swinir} on ×4 upscaling task shown in Fig.~\ref{visual_comparison}. 
Transformer-based model Swin-light~\cite{liang2021swinir} is known for its performance in the state-of-the-art, but we can see that it still has some distortion for repeated pattern information in other images, which can also be observed worse in a CNN-based model IMDN~\cite{Hui-IMDN-2019}.

In ``img\_004'', previous models show undesired artifacts such as ringing artifacts in the form of white bands around the black circle and failed to reconstruct the circle properly, while \proposed{} reconstruct the circle well enough for the HR image.
Also, they have suffered from distortion in ``img\_012'' or reconstruction in the unintended direction of patterns in ``img\_025''.
In contrast, Our \proposed{} model reconstructs patterns straight and in the correct orientation, without distortion.
There are many repeats of long and sharp patterns in ``img\_044'' and ``img\_093''. Additionally, we also observe that previous models either recover it more boldly or blurry, but our \proposed{} has the powerful ability to distinguish them as thin and clear ones.
These comparisons indicate that \proposed{} has a remarkable ability to recover visible details well.

% \FloatBarrier

\section{Discussion on Future Work}

In \proposed{}, the choice of window size significantly impacts both computational efficiency and high-resolution reconstruction performance. Since the optimal window size may vary depending on the pixel distribution of each input image, an adaptive mechanism is needed to dynamically determine the most suitable window size.
As part of our future research, we aim to develop a neural network-based framework that learns to adjust window sizes through fine-tuning on specific datasets, optimizing super-resolution quality accordingly. Additionally, in \proposed{}, the top-k most relevant \fgca{} windows are selected for attention computation. However, the optimal k values may also vary based on the image's internal pixel distribution.
In future work, we plan to explore a type of self-adaptive strategy to determine the optimal k dynamically, ensuring improved performance across diverse image distributions.

\section{Conclusion}
\label{sec:conclusion}
In this paper, we propose \proposed{}, an acceptable framework that introduces query-adaptive context-aware attention to simultaneously capture both local and global features.
Our method selects the top-k most similar key-value windows for each query, ensuring that only the most semantically relevant features contribute to reconstruction while preventing interference from irrelevant regions. To enhance efficiency in resource-constrained environments, we integrate (1) fixed-sized windows and (2) flash attention, enforcing linearity in token-to-token attention while significantly reducing memory usage. Moreover, \proposed{} mitigates the quadratic growth in computation and memory consumption for large-scale images, making it more scalable than existing approaches.
Extensive experiments demonstrate that \proposed{} outperforms recent SISR models across multiple benchmark datasets, highlighting its effectiveness in high-quality super-resolution.

\section*{Acknowledgment}
We would like to express our gratitude to Mr. Jason for his insightful feedback and discussions that facilitated this research. This research was carried out independently, without any external funding or financial support. Instead, we acknowledge the contributions of the open-source community, whose tools and datasets played a crucial role in this research.
{
    \small
    \bibliographystyle{ieeenat_fullname}
    \bibliography{main}

\begin{thebibliography}{55}
\providecommand{\natexlab}[1]{#1}
\providecommand{\url}[1]{\texttt{#1}}
\expandafter\ifx\csname urlstyle\endcsname\relax
  \providecommand{\doi}[1]{doi: #1}\else
  \providecommand{\doi}{doi: \begingroup \urlstyle{rm}\Url}\fi

\bibitem[Aakerberg et~al.(2022)Aakerberg, Nasrollahi, and Moeslund]{aakerberg2022real}
Andreas Aakerberg, Kamal Nasrollahi, and Thomas~B Moeslund.
\newblock Real-world super-resolution of face-images from surveillance cameras.
\newblock \emph{IET Image Processing}, 16\penalty0 (2):\penalty0 442--452, 2022.

\bibitem[Agustsson and Timofte(2017)]{agustsson2017ntire}
Eirikur Agustsson and Radu Timofte.
\newblock Ntire 2017 challenge on single image super-resolution: Dataset and study.
\newblock In \emph{Proceedings of the IEEE conference on computer vision and pattern recognition workshops}, pages 126--135, 2017.

\bibitem[Ahn et~al.(2018)Ahn, Kang, and Sohn]{ahn2018fast}
Namhyuk Ahn, Byungkon Kang, and Kyung-Ah Sohn.
\newblock Fast, accurate, and lightweight super-resolution with cascading residual network, 2018.

\bibitem[Anwar et~al.(2020)Anwar, Khan, and Barnes]{anwar2020deep}
Saeed Anwar, Salman Khan, and Nick Barnes.
\newblock A deep journey into super-resolution: A survey.
\newblock \emph{ACM Computing Surveys (CSUR)}, 53\penalty0 (3):\penalty0 1--34, 2020.

\bibitem[Ayazoglu(2021)]{ayazoglu2021extremely}
Mustafa Ayazoglu.
\newblock Extremely lightweight quantization robust real-time single-image super resolution for mobile devices.
\newblock In \emph{Proceedings of the IEEE/CVF conference on computer vision and pattern recognition}, pages 2472--2479, 2021.

\bibitem[Bevilacqua et~al.(2012)Bevilacqua, Roumy, Guillemot, and Alberi{-}Morel]{bevilacqua2012low}
Marco Bevilacqua, Aline Roumy, Christine Guillemot, and Marie{-}Line Alberi{-}Morel.
\newblock Low-complexity single-image super-resolution based on nonnegative neighbor embedding.
\newblock In \emph{British Machine Vision Conference, {BMVC} 2012, Surrey, UK, September 3-7, 2012}, pages 1--10. {BMVA} Press, 2012.

\bibitem[Cabrera et~al.(2021)Cabrera, Hitefield, Kim, Lee, Miniskar, and Vetter]{cabrera2021toward}
Anthony Cabrera, Seth Hitefield, Jungwon Kim, Seyong Lee, Narasinga~Rao Miniskar, and Jeffrey~S Vetter.
\newblock Toward performance portable programming for heterogeneous systems on a chip: A case study with qualcomm snapdragon soc.
\newblock In \emph{2021 IEEE High Performance Extreme Computing Conference (HPEC)}, pages 1--7. IEEE, 2021.

\bibitem[Chen et~al.(2021)Chen, Wang, Guo, Xu, Deng, Liu, Ma, Xu, Xu, and Gao]{chen2020pre}
Hanting Chen, Yunhe Wang, Tianyu Guo, Chang Xu, Yiping Deng, Zhenhua Liu, Siwei Ma, Chunjing Xu, Chao Xu, and Wen Gao.
\newblock Pre-trained image processing transformer, 2021.

\bibitem[Chen et~al.(2023)Chen, Wang, Zhou, Qiao, and Dong]{chen2023activating}
Xiangyu Chen, Xintao Wang, Jiantao Zhou, Yu Qiao, and Chao Dong.
\newblock Activating more pixels in image super-resolution transformer.
\newblock In \emph{Proceedings of the IEEE/CVF Conference on Computer Vision and Pattern Recognition (CVPR)}, pages 22367--22377, 2023.

\bibitem[Conde et~al.(2023)Conde, Zamfir, Timofte, Motilla, Liu, Zhang, Peng, Lin, Guo, Zou, et~al.]{conde2023efficient}
Marcos~V Conde, Eduard Zamfir, Radu Timofte, Daniel Motilla, Cen Liu, Zexin Zhang, Yunbo Peng, Yue Lin, Jiaming Guo, Xueyi Zou, et~al.
\newblock Efficient deep models for real-time 4k image super-resolution. ntire 2023 benchmark and report.
\newblock In \emph{Proceedings of the IEEE/CVF Conference on Computer Vision and Pattern Recognition}, pages 1495--1521, 2023.

\bibitem[Dao et~al.(2022)Dao, Fu, Ermon, Rudra, and Ré]{dao2022flashattention}
Tri Dao, Daniel~Y. Fu, Stefano Ermon, Atri Rudra, and Christopher Ré.
\newblock Flashattention: Fast and memory-efficient exact attention with io-awareness, 2022.

\bibitem[Dong et~al.(2014)Dong, Loy, He, and Tang]{dong2014learning}
Chao Dong, Chen~Change Loy, Kaiming He, and Xiaoou Tang.
\newblock Learning a deep convolutional network for image super-resolution.
\newblock In \emph{Computer Vision--ECCV 2014: 13th European Conference, Zurich, Switzerland, September 6-12, 2014, Proceedings, Part IV 13}, pages 184--199. Springer, 2014.

\bibitem[Dong et~al.(2016)Dong, Loy, and Tang]{dong2016accelerating}
Chao Dong, Chen~Change Loy, and Xiaoou Tang.
\newblock Accelerating the super-resolution convolutional neural network.
\newblock In \emph{Computer Vision--ECCV 2016: 14th European Conference, Amsterdam, The Netherlands, October 11-14, 2016, Proceedings, Part II 14}, pages 391--407. Springer, 2016.

\bibitem[Dong et~al.(2022)Dong, Bao, Chen, Zhang, Yu, Yuan, Chen, and Guo]{dong2022cswin}
Xiaoyi Dong, Jianmin Bao, Dongdong Chen, Weiming Zhang, Nenghai Yu, Lu Yuan, Dong Chen, and Baining Guo.
\newblock Cswin transformer: A general vision transformer backbone with cross-shaped windows.
\newblock In \emph{Proceedings of the IEEE/CVF conference on computer vision and pattern recognition}, pages 12124--12134, 2022.

\bibitem[Dosovitskiy et~al.(2020)Dosovitskiy, Beyer, Kolesnikov, Weissenborn, Zhai, Unterthiner, Dehghani, Minderer, Heigold, Gelly, et~al.]{dosovitskiy2020image}
Alexey Dosovitskiy, Lucas Beyer, Alexander Kolesnikov, Dirk Weissenborn, Xiaohua Zhai, Thomas Unterthiner, Mostafa Dehghani, Matthias Minderer, Georg Heigold, Sylvain Gelly, et~al.
\newblock An image is worth 16x16 words: Transformers for image recognition at scale.
\newblock \emph{arXiv preprint arXiv:2010.11929}, 2020.

\bibitem[Fu et~al.(2023)Fu, Jiang, Wu, Yan, Wang, and Wang]{fu2023image}
Lihua Fu, Hanxu Jiang, Huixian Wu, Shaoxing Yan, Junxiang Wang, and Dan Wang.
\newblock Image super-resolution reconstruction based on instance spatial feature modulation and feedback mechanism.
\newblock \emph{Applied Intelligence}, 53\penalty0 (1):\penalty0 601--615, 2023.

\bibitem[Gankhuyag et~al.(2023)Gankhuyag, Yoon, Park, Son, and Min]{gankhuyag2023lightweight}
Ganzorig Gankhuyag, Kihwan Yoon, Jinman Park, Haeng~Seon Son, and Kyoungwon Min.
\newblock Lightweight real-time image super-resolution network for 4k images.
\newblock In \emph{Proceedings of the IEEE/CVF Conference on Computer Vision and Pattern Recognition}, pages 1746--1755, 2023.

\bibitem[Hang et~al.(2020)Hang, Liao, Yang, Chen, and Zhou]{hang2020attention}
Yucheng Hang, Qingmin Liao, Wenming Yang, Yupeng Chen, and Jie Zhou.
\newblock Attention cube network for image restoration.
\newblock In \emph{Proceedings of the 28th ACM International Conference on Multimedia}, pages 2562--2570, 2020.

\bibitem[Huang et~al.(2015)Huang, Singh, and Ahuja]{Huang-CVPR-2015}
Jia-Bin Huang, Abhishek Singh, and Narendra Ahuja.
\newblock Single image super-resolution from transformed self-exemplars.
\newblock In \emph{Proceedings of the IEEE Conference on Computer Vision and Pattern Recognition}, pages 5197--5206, 2015.

\bibitem[Hui et~al.(2019{\natexlab{a}})Hui, Gao, Yang, and Wang]{Hui-IMDN-2019}
Zheng Hui, Xinbo Gao, Yunchu Yang, and Xiumei Wang.
\newblock Lightweight image super-resolution with information multi-distillation network.
\newblock In \emph{Proceedings of the 27th ACM International Conference on Multimedia (ACM MM)}, pages 2024--2032, 2019{\natexlab{a}}.

\bibitem[Hui et~al.(2019{\natexlab{b}})Hui, Gao, Yang, and Wang]{hui2019lightweight}
Zheng Hui, Xinbo Gao, Yunchu Yang, and Xiumei Wang.
\newblock Lightweight image super-resolution with information multi-distillation network.
\newblock In \emph{Proceedings of the 27th acm international conference on multimedia}, pages 2024--2032, 2019{\natexlab{b}}.

\bibitem[Khurana et~al.(2023)Khurana, Koli, Khatter, and Singh]{khurana2023natural}
Diksha Khurana, Aditya Koli, Kiran Khatter, and Sukhdev Singh.
\newblock Natural language processing: State of the art, current trends and challenges.
\newblock \emph{Multimedia tools and applications}, 82\penalty0 (3):\penalty0 3713--3744, 2023.

\bibitem[Kim et~al.(2016)Kim, Lee, and Lee]{kim2016accurate}
Jiwon Kim, Jung~Kwon Lee, and Kyoung~Mu Lee.
\newblock Accurate image super-resolution using very deep convolutional networks.
\newblock In \emph{Proceedings of the IEEE conference on computer vision and pattern recognition}, pages 1646--1654, 2016.

\bibitem[Kinli et~al.(2021)Kinli, Ozcan, and Kirac]{kinli2021instagram}
Furkan Kinli, Baris Ozcan, and Furkan Kirac.
\newblock Instagram filter removal on fashionable images.
\newblock In \emph{Proceedings of the IEEE/CVF Conference on Computer Vision and Pattern Recognition}, pages 736--745, 2021.

\bibitem[Ledig et~al.(2017)Ledig, Theis, Husz{\'a}r, Caballero, Cunningham, Acosta, Aitken, Tejani, Totz, Wang, et~al.]{ledig2017photo}
Christian Ledig, Lucas Theis, Ferenc Husz{\'a}r, Jose Caballero, Andrew Cunningham, Alejandro Acosta, Andrew Aitken, Alykhan Tejani, Johannes Totz, Zehan Wang, et~al.
\newblock Photo-realistic single image super-resolution using a generative adversarial network.
\newblock In \emph{Proceedings of the IEEE conference on computer vision and pattern recognition}, pages 4681--4690, 2017.

\bibitem[Lee et~al.(2019)Lee, Venieris, Dudziak, Bhattacharya, and Lane]{lee2019mobisr}
Royson Lee, Stylianos~I Venieris, Lukasz Dudziak, Sourav Bhattacharya, and Nicholas~D Lane.
\newblock Mobisr: Efficient on-device super-resolution through heterogeneous mobile processors.
\newblock In \emph{The 25th annual international conference on mobile computing and networking}, pages 1--16, 2019.

\bibitem[Li et~al.(2021)Li, Zhou, Qi, Jiang, Lu, and Jia]{li2021lapar}
Wenbo Li, Kun Zhou, Lu Qi, Nianjuan Jiang, Jiangbo Lu, and Jiaya Jia.
\newblock Lapar: Linearly-assembled pixel-adaptive regression network for single image super-resolution and beyond, 2021.

\bibitem[Liang et~al.(2021)Liang, Cao, Sun, Zhang, Van~Gool, and Timofte]{liang2021swinir}
Jingyun Liang, Jiezhang Cao, Guolei Sun, Kai Zhang, Luc Van~Gool, and Radu Timofte.
\newblock Swinir: Image restoration using swin transformer.
\newblock In \emph{Proceedings of the IEEE/CVF international conference on computer vision}, pages 1833--1844, 2021.

\bibitem[Lim et~al.(2017)Lim, Son, Kim, Nah, and Mu~Lee]{lim2017enhanced}
Bee Lim, Sanghyun Son, Heewon Kim, Seungjun Nah, and Kyoung Mu~Lee.
\newblock Enhanced deep residual networks for single image super-resolution.
\newblock In \emph{Proceedings of the IEEE conference on computer vision and pattern recognition workshops}, pages 136--144, 2017.

\bibitem[Liu et~al.(2020)Liu, Tang, and Wu]{liu2020residual}
Jie Liu, Jie Tang, and Gangshan Wu.
\newblock Residual feature distillation network for lightweight image super-resolution, 2020.

\bibitem[Liu et~al.(2021{\natexlab{a}})Liu, Li, Fromm, Wang, Jiang, Mariakakis, and Patel]{liu2021splitsr}
Xin Liu, Yuang Li, Josh Fromm, Yuntao Wang, Ziheng Jiang, Alex Mariakakis, and Shwetak Patel.
\newblock Splitsr: An end-to-end approach to super-resolution on mobile devices.
\newblock \emph{Proceedings of the ACM on Interactive, Mobile, Wearable and Ubiquitous Technologies}, 5\penalty0 (1):\penalty0 1--20, 2021{\natexlab{a}}.

\bibitem[Liu et~al.(2021{\natexlab{b}})Liu, Lin, Cao, Hu, Wei, Zhang, Lin, and Guo]{liu2021swin}
Ze Liu, Yutong Lin, Yue Cao, Han Hu, Yixuan Wei, Zheng Zhang, Stephen Lin, and Baining Guo.
\newblock Swin transformer: Hierarchical vision transformer using shifted windows.
\newblock In \emph{Proceedings of the IEEE/CVF international conference on computer vision}, pages 10012--10022, 2021{\natexlab{b}}.

\bibitem[Luo et~al.(2020)Luo, Xie, Zhang, Qu, Li, and Fu]{luo2020latticenet}
Xiaotong Luo, Yuan Xie, Yulun Zhang, Yanyun Qu, Cuihua Li, and Yun Fu.
\newblock Latticenet: Towards lightweight image super-resolution with lattice block.
\newblock In \emph{Computer Vision--ECCV 2020: 16th European Conference, Glasgow, UK, August 23--28, 2020, Proceedings, Part XXII 16}, pages 272--289. Springer, 2020.

\bibitem[Martin et~al.(2001)Martin, Fowlkes, Tal, and Malik]{937655}
D. Martin, C. Fowlkes, D. Tal, and J. Malik.
\newblock A database of human segmented natural images and its application to evaluating segmentation algorithms and measuring ecological statistics.
\newblock In \emph{Proceedings Eighth IEEE International Conference on Computer Vision. ICCV 2001}, pages 416--423 vol.2, 2001.

\bibitem[Mei et~al.(2024)Mei, Hu, Ye, Tang, Wang, Li, Liu, Hao, Lei, Xu, et~al.]{mei2024gtmfuse}
Liye Mei, Xinglong Hu, Zhaoyi Ye, Linfeng Tang, Ying Wang, Di Li, Yan Liu, Xin Hao, Cheng Lei, Chuan Xu, et~al.
\newblock Gtmfuse: Group-attention transformer-driven multiscale dense feature-enhanced network for infrared and visible image fusion.
\newblock \emph{Knowledge-Based Systems}, 293:\penalty0 111658, 2024.

\bibitem[Mei et~al.(2021)Mei, Fan, and Zhou]{Mei_2021_CVPR}
Yiqun Mei, Yuchen Fan, and Yuqian Zhou.
\newblock Image super-resolution with non-local sparse attention.
\newblock In \emph{Proceedings of the IEEE/CVF Conference on Computer Vision and Pattern Recognition (CVPR)}, pages 3517--3526, 2021.

\bibitem[Pathak et~al.(2018)Pathak, Li, Minaee, and Cowan]{pathak2018efficient}
Harsh~Nilesh Pathak, Xinxin Li, Shervin Minaee, and Brooke Cowan.
\newblock Efficient super resolution for large-scale images using attentional gan.
\newblock In \emph{2018 IEEE International Conference on Big Data (Big Data)}, pages 1777--1786. IEEE, 2018.

\bibitem[Rakotonirina and Rasoanaivo(2020)]{rakotonirina2020esrgan+}
Nathana{\"e}l~Carraz Rakotonirina and Andry Rasoanaivo.
\newblock Esrgan+: Further improving enhanced super-resolution generative adversarial network.
\newblock In \emph{ICASSP 2020-2020 IEEE International Conference on Acoustics, Speech and Signal Processing (ICASSP)}, pages 3637--3641. IEEE, 2020.

\bibitem[Tang et~al.(2022)Tang, Zhang, Zhu, and Tan]{tang2022quadtree}
Shitao Tang, Jiahui Zhang, Siyu Zhu, and Ping Tan.
\newblock Quadtree attention for vision transformers.
\newblock \emph{arXiv preprint arXiv:2201.02767}, 2022.

\bibitem[Tian et~al.(2022)Tian, Zhang, Lin, Zuo, Zhang, and Lin]{tian2022generative}
Chunwei Tian, Xuanyu Zhang, Jerry Chun-Wei Lin, Wangmeng Zuo, Yanning Zhang, and Chia-Wen Lin.
\newblock Generative adversarial networks for image super-resolution: A survey.
\newblock \emph{arXiv preprint arXiv:2204.13620}, 2022.

\bibitem[Tu et~al.(2022)Tu, Talebi, Zhang, Yang, Milanfar, Bovik, and Li]{tu2022maxvit}
Zhengzhong Tu, Hossein Talebi, Han Zhang, Feng Yang, Peyman Milanfar, Alan Bovik, and Yinxiao Li.
\newblock Maxvit: Multi-axis vision transformer.
\newblock In \emph{European conference on computer vision}, pages 459--479. Springer, 2022.

\bibitem[Vaswani et~al.(2017)Vaswani, Shazeer, Parmar, Uszkoreit, Jones, Gomez, Kaiser, and Polosukhin]{vaswani2017attention}
Ashish Vaswani, Noam Shazeer, Niki Parmar, Jakob Uszkoreit, Llion Jones, Aidan~N Gomez, {\L}ukasz Kaiser, and Illia Polosukhin.
\newblock Attention is all you need.
\newblock \emph{Advances in neural information processing systems}, 30, 2017.

\bibitem[Wang et~al.(2019)Wang, Li, and Shi]{wang2019lightweight}
Chaofeng Wang, Zhen Li, and Jun Shi.
\newblock Lightweight image super-resolution with adaptive weighted learning network.
\newblock \emph{arXiv preprint arXiv:1904.02358}, 2019.

\bibitem[Wang et~al.(2023)Wang, Chen, Qiu, Chen, Wu, Lin, He, and Liu]{wang2023crossformer}
Wenxiao Wang, Wei Chen, Qibo Qiu, Long Chen, Boxi Wu, Binbin Lin, Xiaofei He, and Wei Liu.
\newblock Crossformer++: A versatile vision transformer hinging on cross-scale attention, 2023.

\bibitem[Wang et~al.(2018)Wang, Yu, Wu, Gu, Liu, Dong, Qiao, and Change~Loy]{wang2018esrgan}
Xintao Wang, Ke Yu, Shixiang Wu, Jinjin Gu, Yihao Liu, Chao Dong, Yu Qiao, and Chen Change~Loy.
\newblock Esrgan: Enhanced super-resolution generative adversarial networks.
\newblock In \emph{Proceedings of the European conference on computer vision (ECCV) workshops}, pages 0--0, 2018.

\bibitem[Wang et~al.(2020)Wang, Wang, Zhao, Yan, Fan, and Chen]{wang2020lightweight}
Xuehui Wang, Qing Wang, Yuzhi Zhao, Junchi Yan, Lei Fan, and Long Chen.
\newblock Lightweight single-image super-resolution network with attentive auxiliary feature learning, 2020.

\bibitem[Wang et~al.(2021)Wang, Huang, Song, Huang, and Huang]{wang2021not}
Yulin Wang, Rui Huang, Shiji Song, Zeyi Huang, and Gao Huang.
\newblock Not all images are worth 16x16 words: Dynamic transformers for efficient image recognition.
\newblock \emph{Advances in neural information processing systems}, 34:\penalty0 11960--11973, 2021.

\bibitem[Xie et~al.(2022)Xie, Zhang, Xia, Hengel, and Wu]{xie2022clustr}
Yutong Xie, Jianpeng Zhang, Yong Xia, Anton van~den Hengel, and Qi Wu.
\newblock Clustr: Exploring efficient self-attention via clustering for vision transformers.
\newblock \emph{arXiv preprint arXiv:2208.13138}, 2022.

\bibitem[Xu et~al.(2024)Xu, Park, Zhang, Zhou, Shechtman, Liu, Huang, and Liu]{xu2024videogigagan}
Yiran Xu, Taesung Park, Richard Zhang, Yang Zhou, Eli Shechtman, Feng Liu, Jia-Bin Huang, and Difan Liu.
\newblock Videogigagan: Towards detail-rich video super-resolution.
\newblock \emph{arXiv preprint arXiv:2404.12388}, 2024.

\bibitem[Yang et~al.(2021)Yang, Li, Zhang, Dai, Xiao, Yuan, and Gao]{yang2021focal}
Jianwei Yang, Chunyuan Li, Pengchuan Zhang, Xiyang Dai, Bin Xiao, Lu Yuan, and Jianfeng Gao.
\newblock Focal self-attention for local-global interactions in vision transformers.
\newblock \emph{arXiv preprint arXiv:2107.00641}, 2021.

\bibitem[Zeyde et~al.(2012)Zeyde, Elad, and Protter]{10.1007/978-3-642-27413-8_47}
Roman Zeyde, Michael Elad, and Matan Protter.
\newblock On single image scale-up using sparse-representations.
\newblock In \emph{Curves and Surfaces}, pages 711--730, Berlin, Heidelberg, 2012. Springer Berlin Heidelberg.

\bibitem[Zhang et~al.(2022{\natexlab{a}})Zhang, Huang, Liu, Wang, and Jin]{zhang2022swinfir}
Dafeng Zhang, Feiyu Huang, Shizhuo Liu, Xiaobing Wang, and Zhezhu Jin.
\newblock Swinfir: Revisiting the swinir with fast fourier convolution and improved training for image super-resolution.
\newblock \emph{arXiv preprint arXiv:2208.11247}, 2022{\natexlab{a}}.

\bibitem[Zhang et~al.(2024)Zhang, Kasichainula, Zhuo, Li, Seo, and Cao]{zhang2024transformer}
Tianyi Zhang, Kishore Kasichainula, Yaoxin Zhuo, Baoxin Li, Jae-Sun Seo, and Yu Cao.
\newblock Transformer-based selective super-resolution for efficient image refinement.
\newblock In \emph{Proceedings of the AAAI Conference on Artificial Intelligence}, pages 7305--7313, 2024.

\bibitem[Zhang et~al.(2022{\natexlab{b}})Zhang, Zeng, Guo, and Zhang]{zhang2022efficient}
Xindong Zhang, Hui Zeng, Shi Guo, and Lei Zhang.
\newblock Efficient long-range attention network for image super-resolution, 2022{\natexlab{b}}.

\bibitem[Zhu et~al.(2023)Zhu, Wang, Ke, Zhang, and Lau]{zhu2023biformer}
Lei Zhu, Xinjiang Wang, Zhanghan Ke, Wayne Zhang, and Rynson~WH Lau.
\newblock Biformer: Vision transformer with bi-level routing attention.
\newblock In \emph{Proceedings of the IEEE/CVF Conference on Computer Vision and Pattern Recognition}, pages 10323--10333, 2023.

\end{thebibliography}
}


\end{document}